%% file: template.tex
\documentclass[journal]{vgtc}                     

\onlineid{0}

\vgtccategory{Research}

\vgtcpapertype{please specify}

\title{VRGaussianAvatar: Integrating 3D Gaussian Avatars into VR}

\author{%
  \authororcid{Hail Song}{0009-0006-4008-196X},
  \authororcid{Boram Yoon}{0000-0003-3696-0145},
  \authororcid{Seokhwan Yang}{0009-0000-1097-646X},
  \authororcid{Seoyoung Kang}{0000-0002-6143-4369},\\
  \authororcid{Hyunjeong Kim}{0000-0003-0156-5436},
  \authororcid{Henning Metzmacher}{0000-0002-1120-9534}, and
  \authororcid{Woontack Woo}{0000-0002-5501-4421}
}

\authorfooter{
  \item
    Hail Song is with KAIST UVR Lab.
    E-mail: hail96@kaist.ac.kr
  \item
    Boram Yoon is with KAIST KI-ITC ARRC.
    E-mail: boram.yoon1206@kaist.ac.kr
  \item
    Seokhwan Yang is with KAIST UVR Lab.
    E-mail: ysshwan147@kaist.ac.kr
  \item
    Seoyoung Kang is with KAIST UVR Lab.
    E-mail: sy1009kang@kaist.ac.kr
  \item
    Hyunjeong Kim is with Hansung University and ETH Zurich, Game Technology Center.
    E-mail: hcihjkim@gmail.com
  \item
    Henning Metzmacher is with ETH Zurich, Game Technology Center.
    E-mail: henning.metzmacher@inf.ethz.ch
  \item
    Woontack Woo is with KAIST UVR Lab and KAIST KI-ITC ARRC.
    E-mail: wwoo@kaist.ac.kr

}

\abstract{%
We present VRGaussianAvatar, an integrated system that enables real-time full-body 3D Gaussian Splatting (3DGS) avatars in virtual reality using only head-mounted display (HMD) tracking signals. The system adopts a parallel pipeline with a VR Frontend and a GA Backend. The VR Frontend uses inverse kinematics to estimate full-body pose and streams the resulting pose along with stereo camera parameters to the backend. The GA Backend stereoscopically renders a 3DGS avatar reconstructed from a single image. To improve stereo rendering efficiency, we introduce Binocular Batching, which jointly processes left and right eye views in a single batched pass to reduce redundant computation and support high-resolution VR displays. We evaluate VRGaussianAvatar with quantitative performance tests and a within-subject user study against image- and video-based mesh avatar baselines. Results show that VRGaussianAvatar sustains interactive VR performance and yields higher perceived appearance similarity, embodiment, and plausibility. Project page and source code are available at \url{https://vrgaussianavatar.github.io}
  
}

\teaser{
  \centering
  \includegraphics[width=0.93\linewidth]{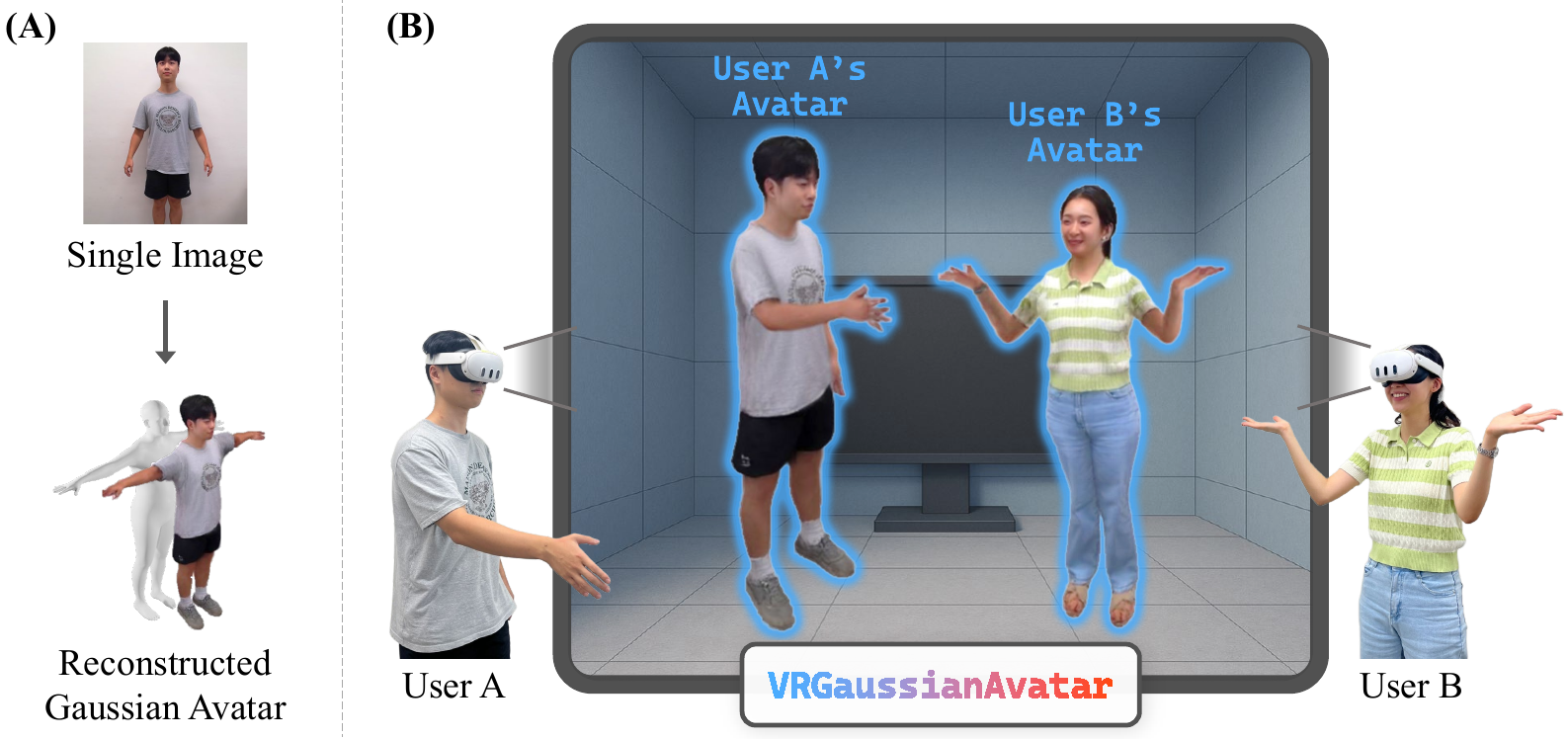}
\caption{
We introduce \textbf{VRGaussianAvatar}, an integrated VR system for controllable full-body 3DGS avatar representation. 
(A) A single input image is used to reconstruct a full-body 3D Gaussian Splatting avatar. 
(B) The reconstructed avatar supports real-time rendering and full controllability in VR using only the internal sensors of commercial head-mounted displays (HMDs).
}
  \label{fig:teaser}
}


\graphicspath{{figs/}{figures/}{pictures/}{images/}{./}} 

\usepackage{tabu}                      
\usepackage{booktabs}                  
\usepackage{lipsum}                    
\usepackage{mwe}                       

\usepackage{amssymb}
\usepackage{mathptmx}                  
\usepackage{booktabs}
\usepackage{multirow}
\usepackage{amsmath}
\usepackage[inkscapelatex=false]{svg}
\usepackage{enumitem}
\setlist[itemize]{topsep=2pt, itemsep=2pt, parsep=0pt}
\setlist[enumerate]{topsep=2pt, itemsep=2pt, parsep=0pt}
\usepackage{makecell}

\begin{document}



\maketitle

\input{sections_revised/1_introduction}

\input{sections_revised/2_Related_work}

\input{sections_revised/3_0_Method}

\input{sections_revised/4_Experiments}
\input{sections_revised/5_Results}

\input{sections_revised/6_Conclusion}

\acknowledgments{
This work was supported by the National Research Council of Science \& Technology (NST) grant by the Korea government (MSIT) (No. CRC21015).
This research was supported by the MSIT(Ministry of Science and ICT), Korea, under the Graduate School of Metaverse Convergence support program(IITP-2026-RS-2022-00156435) supervised by the IITP(Institute for Information \& Communications Technology Planning \& Evaluation).
This work was supported by the Institute of Information \& Communications Technology Planning \& Evaluation(IITP) grant funded by the Korea government(MSIT) (No. RS-2025-25441313, Professional AI Talent Development Program for Multimodal AI Agents).
This research was financially supported by Hansung University for Hyunjeong Kim.
This work was supported by Institute of Information \& communications Technology Planning \& Evaluation (IITP) grant funded by the Korea government(MSIT) (RS-2022-00143911, AI Excellence Global Innovative Leader Education Program).
}

\bibliographystyle{abbrv-doi-hyperref}

\bibliography{template}

\appendix 

\input{sections_revised/7_Appendix}

\end{document}

%% file: sections_revised/1_introduction.tex
\section{Introduction}
The representation of photorealistic full-body avatars in Virtual Reality (VR) plays a crucial role in enhancing user immersion and effective communication. Applications such as telepresence and gaming benefit significantly from avatars that faithfully replicate the user’s appearance.

To make avatar creation practical and accessible, existing methods allow for the creation of mesh-based 3D avatars. These methods reconstruct drivable human models from a single image~\cite{SMPL-X, Kanazawa_2018_CVPR, feng2021deca, cai2023smpler, yin2025smplest} or from video sequences~\cite{alldieck2018video,rcsmpl,kocabas2020vibe}.
While these approaches are highly compatible with existing graphics pipelines and offer strong real-time performance, they often struggle to deliver photorealistic avatar representations for subjects that significantly deviate from template meshes such as SMPL-X~\cite{SMPL-X}.
Recently, neural rendering techniques, including Neural Radiance Fields (NeRF)~\cite{mildenhall2021nerf} and 3D Gaussian Splatting (3DGS)~\cite{3dgaussiansplatting}, have achieved remarkable success across diverse 3D visualization tasks. These methods have been applied not only to novel view synthesis but also to object reconstruction~\cite{tang2023dreamgaussian, tang2024lgm, lee2024vividdream} and avatar creation~\cite{jiang2022instantavatar, chatziagapi2024talkinnerf, gao2022reconstructing}. In particular, 3DGS represents a 3D scene as a collection of colored Gaussian splats, enabling real-time photorealistic rendering and demonstrating strong performance in both face~\cite{zheng2024headgap, tran2024voodoo3d, he2025lam} and body avatar generation~\cite{moon2024expressive, qiu2025lhm, sim2025persona}.

In VR, several studies~\cite{jiang2024vrgs, tu2025vrsplat} have proposed techniques that leverage 3DGS for immersive scene representation.
Kim et al.\cite{kim2024is3dgs} conducted quantitative evaluations and user studies to examine the effectiveness of 3DGS-based object representation in VR, confirming that 3DGS is effective for realistic object rendering in virtual environments. 
VOODOO XP~\cite{tran2024voodooxp} was the first to propose a VR telepresence application using a 3DGS-based head avatar animated using the internal sensors of head-mounted displays (HMDs).

However, several challenges have prevented 3DGS-based full-body avatars from being deployed in VR communication systems.
In addition to the significant hardware demands of 3DGS visualization, VR requires stereoscopic rendering which imposes even higher computational demands.
Moreover, 3DGS pipelines have limited compatibility with conventional VR rendering pipelines. Real-time animation of 3DGS avatars remains challenging, and existing systems do not directly utilize user motion captured from the internal sensors of HMDs. These limitations have hindered the integration of 3DGS avatars into VR. 

To overcome these limitations, we present \textbf{VRGaussianAvatar}, an integrated system that enables real-time user representation in VR through a 3DGS avatar reconstructed from a single image and controllable using only the internal sensors of a commercial HMD.
Our system consists of two subsystems: a VR Frontend, which processes sensor data from a commercial HMD and applies inverse kinematics for avatar animation, and a GA Backend (Gaussian Avatar Backend), which performs real-time photorealistic rendering with 3DGS.
To support stereoscopic VR rendering, we introduce \textbf{Binocular Batching}, which achieves efficient and consistent rendering for both eyes.
The system builds on standard components (e.g., off-the-shelf inverse kinematics and recent single-image 3DGS avatar reconstruction models), while our main contributions are the VR streaming integration, Binocular Batching, and a perceptual user study of 3DGS avatars in VR.

We validate our system through user studies and quantitative evaluations comparing VRGaussianAvatar with mesh-based avatars reconstructed from images and videos.
These mesh avatars were built upon the parametric whole-body model SMPL-X~\cite{SMPL-X} and were designed to be controlled solely through the internal sensors of the HMD, similar to ours. 
For the user study, subjects followed instructions to control both baseline and our method’s avatars, modeled after themselves and others. They then completed questionnaires assessing virtual embodiment and plausibility.
The results show that our method provides stronger self-embodiment than mesh-based baselines. Quantitative evaluations further confirm that avatars produced by our approach resemble the user’s real appearance more closely.


\noindent In summary, our contributions are as follows:
\begin{itemize}
    \item \textbf{VRGaussianAvatar}: an integrated VR streaming system for stereoscopic rendering of a 3DGS full-body avatar driven by HMD-only tracking.
    \item \textbf{Binocular Batching}: an efficient stereoscopic rendering strategy that jointly renders the left and right eye views in a single batched pass for 3DGS avatars in VR.
    \item A quantitative evaluation and perceptual user study showing improved perceived appearance similarity, embodiment, and plausibility over mesh-based avatar baselines under the same HMD-only control setting.
\end{itemize}

%% file: sections_revised/2_Related_work.tex
\section{Related Work}

\subsection{Avatar Mediated Communication}

In VR environments, virtual embodiment and avatar plausibility during communication are both crucial. For effective VR communication, it is important not only that individuals identify with their own avatars through virtual embodiment, but also that avatars are perceived as plausible in appearance and behavior. Previous studies have shown that visual appearance plays a key role in virtual embodiment and more realistic appearances resembling actual humans significantly enhance embodiment~\cite{weidner2023systematic}.
To achieve high levels of plausibility in VR communication, avatar realism is crucial. Studies have concluded that high-quality avatars enhance plausibility-related judgments and user comfort.
For example, Yoon et al.~\cite{yoon2019effect} demonstrated that realistic full-body avatars are advantageous regarding virtual embodiment and plausibility.
Similarly, Cho et al.~\cite{cho2020effects} showed that realistic volumetric avatars yield higher perceived plausibility than pre-scanned avatars when performing both dynamic and static tasks. 
Moreover, Kaiser et al.~\cite{kaiser2025get} reported through quantitative and qualitative experiments that higher avatar realism generally improves plausibility and comfort.

In this work, we present a VR avatar representation system that leverages a full-body avatar based on 3D Gaussian Splatting (3DGS), which has recently shown strong performance in realistic 3D human body model representation. Furthermore, through user studies and statistical analysis, we confirm that the proposed Gaussian avatar representation system contributes to increasing users’ virtual embodiment and avatar plausibility.

\subsection{3D Avatar Reconstruction}
The generation of 3D avatars resembling users has been a critical topic in the areas of computer vision, computer graphics, and virtual reality. For full-body avatar reconstruction, the parametric body model SMPL-X~\cite{SMPL-X} has been widely adopted, as it encodes body shape and facial expressions into latent features and enables pose-dependent 3D representations and animations driven by body, hand, and facial configurations.
Using SMPL-X as a template mesh, several studies have explored approaches to reconstruct mesh and texture maps from videos~\cite{alldieck2018video, rcsmpl, zhi2020texmesh} or images~\cite{SMPL-X, liu2024texdreamer}.

The advancement of neural rendering techniques, represented by neural radiance fields (NeRF)~\cite{mildenhall2021nerf}, has demonstrated strong performance in generating photorealistic full-body avatars from videos~\cite{jiang2022instantavatar, zheng2023avatarrex, chatziagapi2024talkinnerf, yu2023monohuman} or images~\cite{hu2023sherf}. However, the high computational cost and long processing times of these methods limit their applicability to real-time avatar representation in VR environments. To achieve realistic 3D representations with efficient rendering, 3DGS was introduced as a neural rendering technique designed for both fidelity and speed. 3DGS enables fast and photorealistic avatar representation from videos~\cite{moon2024expressive, qian20233dgsavatar, liu2024gea, peng2025rmavatar, moreau2024human} or image~\cite{Men_2024_CVPR, qiu2025lhm, sim2025persona, qiu2025pf} and has also shown superior performance in head model reconstruction~\cite{tran2024voodoo3d, tran2024voodooxp, zheng2024headgap, song2024toward, gaussianavatars, chen2023monogaussianavatar, zhao2024psavatar} compared to NeRF-based approaches.

In particular, the Large Human Model (LHM)~\cite{qiu2025lhm} achieves one-shot avatar reconstruction in just a few seconds while supporting real-time rendering on consumer-grade GPUs.
We adopt LHM~\cite{qiu2025lhm} as an off-the-shelf backbone to reconstruct 3D Gaussian avatars, and focus on integrating it into a practical VR streaming and stereoscopic rendering system.

\begin{figure*} [ht!]
  \includegraphics[width=\textwidth]{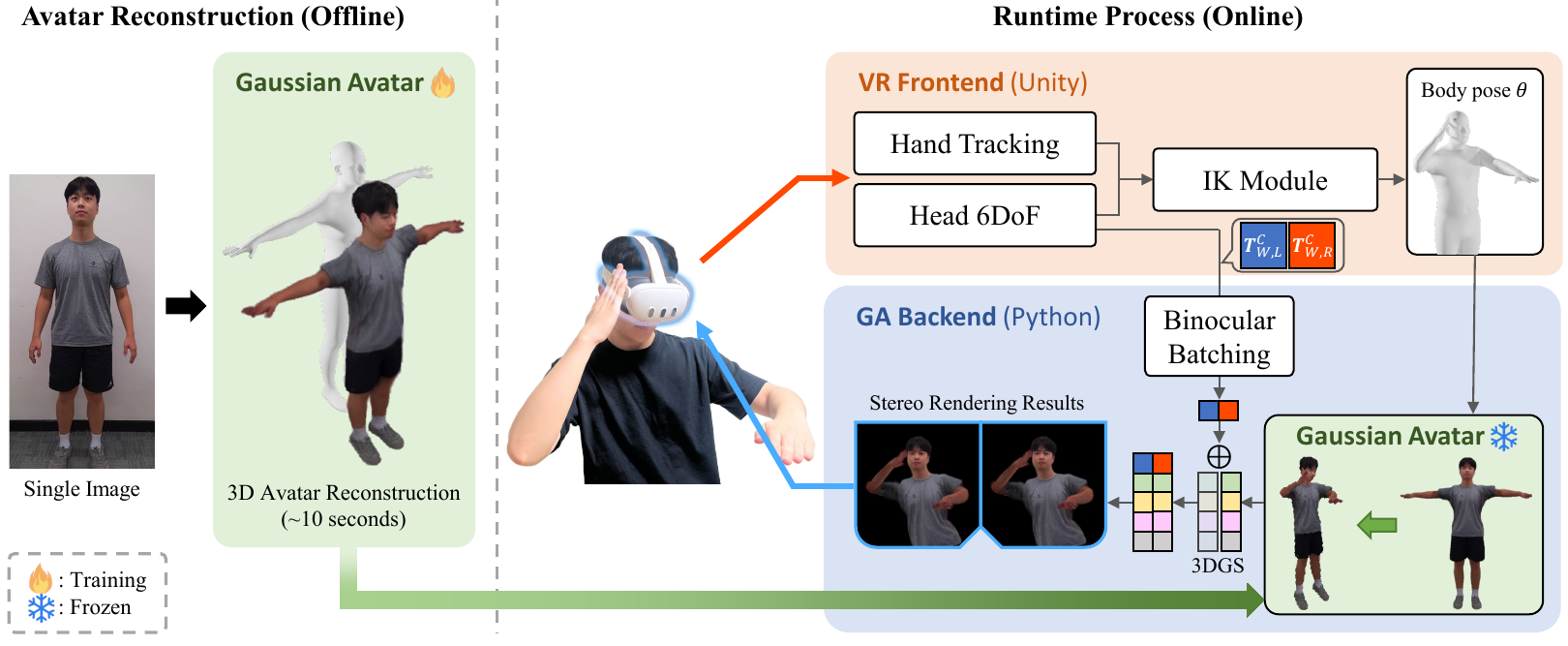}
  \caption{
    System diagram of the proposed method, \textbf{VRGaussianAvatar}. 
    \textit{Avatar reconstruction phase (left):} Given a single input image, the Gaussian Avatar Module reconstructs a 3D full-body avatar. 
    \textit{Runtime process (right):} The \textbf{VR Frontend} and \textbf{GA Backend} operate in parallel to animate and render the avatar in real time. 
    The VR Frontend estimates a SMPL-X–compatible full-body pose from head 6DoF and hand-tracking signals using the IK Module. 
    The GA Backend animates and renders the reconstructed Gaussian avatar, 
    and applies the proposed \textbf{Binocular Batching} with both world-to-camera matrices ($\textbf{T}^C_{W,L}, \textbf{T}^C_{W,R}$) to enable efficient real-time stereoscopic rendering.}

  \label{fig:overview}
\end{figure*}

\subsection{VR Avatar Representation Using Neural Rendering}
There have been several studies that explored the use of neural rendering for photorealistic scene and object representation in VR. Immersive-NGP~\cite{li2022immersive}, which applied the hash-table–based instant-NGP~\cite{muller2022instant} to accelerate NeRF rendering to near real-time, was the first work to demonstrate an application of neural rendering in VR. VR-GS~\cite{jiang2024vrgs} further leveraged 3DGS, which enables both faster rendering and more realistic 3D representations, to achieve real-time VR rendering and introduced physics-based interactions. However, these works have been limited in that they cannot be directly applied to neural rendering–based VR avatar representation, which requires real-time rendering of dynamically moving neural primitives.

To address this, SqueezeMe~\cite{iandola2025squeezeme} reduced the computational overhead of animation and 3DGS rendering, making real-time full-body avatar representation in VR. For head avatars, VOODOO XP~\cite{tran2024voodooxp} demonstrated the animation and rendering of a 3DGS head avatar driven by the internal sensors of HMD. Nevertheless, no prior work has realized a full-body avatar representation using only the limited information from HMD-based body tracking and head tracking.
Motivated by these gaps, we integrate a full-body 3DGS avatar into a VR streaming pipeline and evaluate its perceptual impact in VR.


%% file: sections_revised/3_0_Method.tex
\section{Method}

\textbf{VRGaussianAvatar} is structured as a parallel framework with a \textbf{VR Frontend} and a \textbf{GA Backend}.
The VR Frontend estimates full-body pose from HMD sensors and provides per-frame camera parameters.
The GA Backend animates and renders the pre-reconstructed 3DGS avatar, incorporating a \textbf{Binocular Batching} strategy for efficient stereoscopic rendering.
In our reference setup, the VR Frontend and GA Backend communicate over localhost to minimize network overhead.
\autoref{fig:overview} presents the overall system architecture.

\input{sections_revised/3_1_VR_Frontend}

\input{sections_revised/3_2_GA_Backend}

%% file: sections_revised/3_1_VR_Frontend.tex
\subsection{VR Frontend}
\label{subsec:frontend}

\textbf{Inputs and Objective.\space\space}
The VR Frontend uses only device-native tracking signals from a commercial HMD: head and hand poses (position and orientation).
Its goal is to estimate a full-body pose at interactive rates from these sparse cues and to stream per-frame stereo camera parameters to the GA Backend.

~\\
\textbf{Inverse Kinematics Module (IK Module).\space\space}
The VR Frontend reconstructs a full-body pose using an off-the-shelf IK solver.
The head pose provides the root constraint via a calibrated head-to-root offset.
When available, left/right hand tracking is used as end-effector constraints to stabilize upper-limb articulation; otherwise, the solver falls back to a head-only configuration with posture priors on the torso and shoulders.
The solver enforces joint limits and applies temporal smoothing for stability.
The output is a full-body pose parameter set $\theta$ (SMPL-X compatible) and the head pose $\mathbf{T}^{W}_{H}$ used to derive stereo cameras.

~\\
\textbf{Stereo Camera Construction.\space\space}
From the head pose, we construct per-eye camera transforms and transmit them to the GA Backend.
We place an interpupillary baseline $\mathrm{IPD}$ along the head local $+X$ axis:
\begin{equation}
\label{eq:eye_extrinsics}
\mathbf{T}^{H}_{L}=
\begin{bmatrix}
\mathbf{I} & (-\mathrm{IPD}/2,\,0,\,0)^\top\\
\mathbf{0}^\top & 1
\end{bmatrix},\qquad
\mathbf{T}^{H}_{R}=
\begin{bmatrix}
\mathbf{I} & (\mathrm{IPD}/2,\,0,\,0)^\top\\
\mathbf{0}^\top & 1
\end{bmatrix}.
\end{equation}
World-space eye poses are obtained by composition:
\begin{equation}
\label{eq:world_eye_pose}
\mathbf{T}^{W}_{e}=\mathbf{T}^{W}_{H}\,\mathbf{T}^{H}_{e}, \qquad e\in\{L,R\}.
\end{equation}
The GA Backend uses the corresponding world-to-camera transforms:
\begin{equation}
\label{eq:world_to_camera}
\mathbf{T}^{C}_{W,e}=(\mathbf{T}^{W}_{e})^{-1}.
\end{equation}

We use the standard pinhole camera model to derive per-eye intrinsics and projection from the headset-provided field of view and render resolution.

~\\
\textbf{Per-Frame Packaging.\space\space}
For each frame, the VR Frontend sends a compact request
$\{\texttt{frame\_id},(\mathbf{T}^{C}_{W,e},\mathbf{K}),\theta\}$ to the GA Backend,
where \texttt{frame\_id} is monotonically increasing, $e\in\{L,R\}$, $\mathbf{K}$ denotes shared camera intrinsics, and $\theta$ is the estimated body pose.

~\\
\textbf{Implementation Details.\space\space}
In our reference implementation, the VR Frontend is developed with Unity 6000.1.13f1\footnote{\url{https://unity.com}}.
The IK Module employs Unity Final IK\footnote{\url{https://assetstore.unity.com/packages/tools/animation/final-ik-14290}}.
We use the Meta Quest~3\footnote{\url{https://store.facebook.com/quest/products/quest-3}} as the HMD.

%% file: sections_revised/3_2_GA_Backend.tex
\subsection{GA Backend}
\label{subsec:backend}
\textbf{Role and Interface.\space\space}
The GA Backend receives per-frame requests from the VR Frontend and renders a 3DGS avatar reconstructed from a single image. Each request includes a sequential \texttt{frame\_id}, per-eye camera parameters for $L$ and $R$ (world-to-camera transforms $\mathbf{T}^{C}_{W,e}\!\in\!\mathrm{SE}(3)$ for $e\!\in\!\{L,R\}$ and a shared intrinsic matrix $\mathbf{K}$), and a full-body pose estimate. The backend returns a stereo pair that is streamed directly to the HMD.


~\\
\textbf{Binocular Batching.\space\space}
We introduce Binocular Batching, which processes the two eye views of a frame within a single batch over a shared 3DGS state.
In contrast to a naive stereo pipeline that renders the left and right eye sequentially with duplicated computation, our method stages per-Gaussian attributes $(p_i,\sigma_i,r_i,\rho_i,f_i)$ only once in GPU memory and reuses them across both eyes. 
Eye-specific projection and visibility $(\mathbf{K},\mathbf{T}^{C}_{W,e})$ are then evaluated separately within the same pass. 
This reduces redundant memory transfers and per-frame fixed overhead (e.g., kernel launches, data marshalling, cache warm-up), while improving cache coherence and memory locality.
As a result, both eye buffers are produced in a single batched pass, yielding a notable reduction in rendering time, which we analyze in detail in \autoref{sec:5.2}.

~\\
\textbf{Gaussian Avatar Module.\space\space}
The Gaussian Avatar Module estimates canonical 3D Gaussian parameters, SMPL-X rig weights, and animation-ready bindings from a single image.
In our implementation, this module is instantiated using the Large Animatable Human Model (LHM)~\cite{qiu2025lhm} as an off-the-shelf reconstructor.
The rest of the GA Backend is agnostic to the reconstructor, and any method that outputs $(\chi,\{w_{ik}\},\{T_k(\cdot)\})$ can be plugged in.


~\\
\textit{Tokenization} : 
We encode SMPL\mbox{-}X geometry and image appearance into transformer tokens. 
Let $X=\{x_j\}_{j=1}^{N_{\text{pts}}}\!\subset\!\mathbb{R}^3$ be canonical SMPL\mbox{-}X surface samples. 
Here, $N_{\text{pts}}$ denotes the number of sampled surface points, which we set to 2,048 in our implementation.
After a sinusoidal positional encoding $\gamma:\mathbb{R}^3\!\to\!\mathbb{R}^{3L}$ and a modality-specific projection to the channel dimension $C$, we obtain
\begin{equation}
T^{3D} \;=\; \mathrm{MLP}_{\text{proj}}\!\big(\gamma(X)\big) \;\in\; \mathbb{R}^{N_{\text{pts}}\times C}.
\end{equation}
From the input image $I\!\in\!\mathbb{R}^{H\times W\times 3}$, a frozen body encoder $E_{\text{body}}$ produces features in $\mathbb{R}^{N_{\text{body}}\times C_b}$ which are projected to
\begin{equation}
T^{\text{body}} \;=\; \mathrm{MLP}_{\text{proj}}\!\big(E_{\text{body}}(I)\big) \;\in\; \mathbb{R}^{N_{\text{body}}\times C}.
\end{equation}
To retain fine facial details, a frozen multi-scale head encoder yields $\{E_{\text{head}}^{(d)}(I)\in\mathbb{R}^{N_{\text{head}}^{(d)}\times C_{h,d}}\}_{d=1}^{D}$; a fusion module aggregates them into
\begin{equation}
T^{\text{head}} \;=\; F_{\text{fusion}}\!\Big(\{\,E_{\text{head}}^{(d)}(I)\,\}_{d=1}^{D}\Big) \;\in\; \mathbb{R}^{N_{\text{head}}\times C},
\end{equation}
where $N_{\text{head}}$ denotes the post-fusion token count.

~\\
\textit{Prediction} : 
A Multimodal Body–Head Transformer (MBHT) fuses the three token sets and a decoder predicts canonical Gaussians and rig parameters:
\begin{equation}
\big(\chi,\;\{w_{ik}\},\;\{T_k(\cdot)\}\big)
\;=\;
\mathrm{Decoder}\!\Big(\mathrm{MBHT}\big(T^{3D},\,T^{\text{body}},\,T^{\text{head}}\big)\Big).
\end{equation}
Here, $\chi$ contains $N$ Gaussians, $w_{ik}\!\ge\!0$ with $\sum_k w_{ik}=1$, and $T_k(\theta)\!\in\!\mathrm{SE}(3)$ are rigid bone transforms defined in the canonical space.

Concretely, the canonical Gaussian set is
\begin{equation}
\chi \;=\; \Big\{(p_i,\ \sigma_i,\ r_i,\ \rho_i,\ f_i)\Big\}_{i=1}^{N},
\end{equation}
where $p_i\!\in\!\mathbb{R}^3$ (centroid), $\sigma_i\!\in\!\mathbb{R}^3$ (anisotropic scales), $r_i\!\in\!\mathbb{S}^3$ (unit quaternion), $\rho_i\!\in[0,1]$ (opacity), and $f_i\!\in\!\mathbb{R}^C$ (SH appearance). The 3D covariance is
\begin{equation}
\Sigma_i \;=\; R(r_i)\,\mathrm{diag}(\sigma_i^2)\,R(r_i)^\top,\qquad R(r_i)\in\mathrm{SO}(3),
\end{equation}
with $R(\cdot)$ the quaternion-to-rotation map. 
To enable animation, the module provides an SMPL\mbox{-}X\!-compatible rig with blend weights $w_{ik}$ and bone transforms $T_k(\theta)\!\in\!\mathrm{SE}(3)$ for pose $\theta$.

~\\
\textit{Pose conditioning (runtime)} :
Given the per-frame pose parameters $\theta$ from the VR Frontend, we deform the canonical Gaussian set $\chi$ into a posed set $\chi'=\mathrm{LBS}(\chi,\theta)$ using linear blend skinning (LBS) with per-Gaussian weights $w_{ik}$.
We define the LBS output for Gaussian $i$ as $(p_i', \Sigma_i')=\mathrm{LBS}(p_i,\Sigma_i;\theta)$.
Specifically, letting $T_k(\theta)=(R_k(\theta),t_k(\theta))$ denote the pose-dependent transform of bone $k$ and $R_i(\theta)$ the induced blended rotation, we have
\begin{align}
p_i' &= \sum_{k} w_{ik}\,\big(R_k(\theta)\,p_i + t_k(\theta)\big), \\
\Sigma_i' &= R_i(\theta)\,\Sigma_i\,R_i(\theta)^\top.
\end{align}

~\\
\textit{Splatting and compositing} : 
We render posed Gaussians using the standard 3D Gaussian Splatting rasterization~\cite{3dgaussiansplatting}.
\begin{equation}
\label{eq:compositing}
\mathbf{I}_e \;=\; \mathrm{Render}\big(\mathrm{LBS}(\chi,\theta),\,\mathbf{K},\,\mathbf{T}^{C}_{W,e}\big),\qquad e\in\{L,R\},
\end{equation}
where $\mathrm{LBS}(\chi,\theta)$ denotes the posed Gaussian set after the LBS step above, and $(\mathbf{K},\mathbf{T}^{C}_{W,e})$ are the per-eye camera parameters.
Under Binocular Batching, per-Gaussian attributes are staged once and reused for both $L$ and $R$ within a single batch, while per-eye projection and view-dependent shading are evaluated per eye in the same pass.

\begin{figure}[t!]
\centering
 \includegraphics[width=0.99\columnwidth]{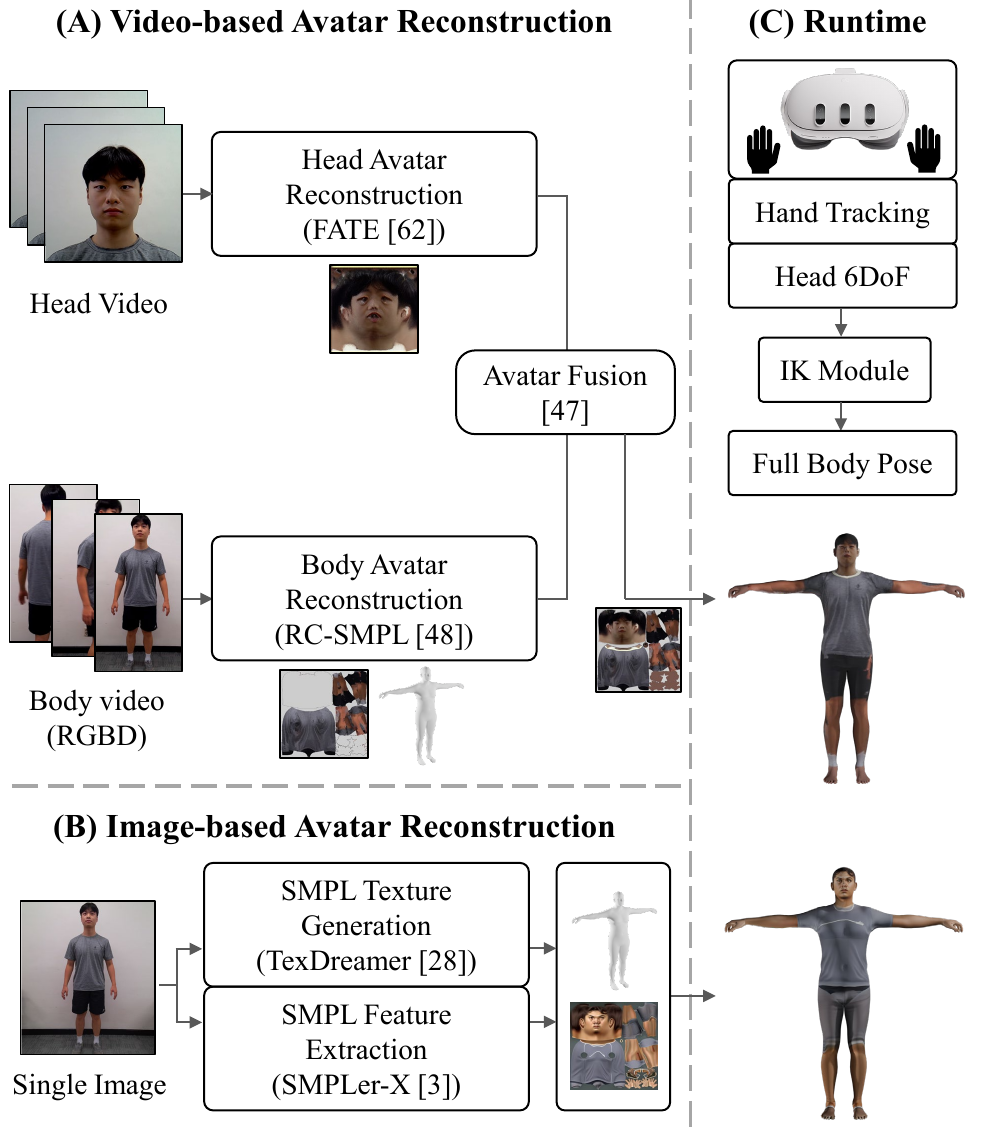}
\caption{
\textbf{Baseline conditions.} 
(A) \textit{video-based reconstruction:} Implemented following prior works~\cite{rcsmpl, zhang2025fate, song2025fasttexturetransferxr}. 
(B) \textit{image-based reconstruction:} Implemented following prior works~\cite{liu2024texdreamer, cai2023smpler}. 
(C) \textit{Runtime process:} The IK Module estimates a full-body pose from head 6DoF and hand-tracking signals to animate the avatar in real time. 
}
\label{fig:baseline}
\end{figure}

~\\
\textbf{Encoding and Delivery to HMD.\space\space}
Rendered frames are converted to 8-bit sRGB and JPEG-compressed at a fixed quality; the left/right payloads, together with \texttt{frame\_id}, are transmitted in a single response and delivered directly to the HMD.
The stateless per-frame design admits deterministic dropping of late frames on the receiver.

The GA Backend renders the 3DGS avatar with an alpha channel to support optional avatar compositing when a 3DGS scene/background is included.
In our study, we disable background compositing and use a black background to minimize background influence and focus the evaluation on the avatar itself; accordingly, we stream RGB-only JPEG frames to reduce bandwidth.

~\\
\textbf{Algorithmic Outline.\space\space}
\begin{enumerate}
\setlength{\itemsep}{1pt}
\setlength{\topsep}{1pt}
  \item Receive $\{\texttt{frame\_id},(\mathbf{T}^{C}_{W,e},\mathbf{K}),\theta\}$ from the VR Frontend.

  \item Construct a single-batch stereo rendering context and apply pose conditioning to the canonical 3DGS avatar, 
  deforming Gaussian primitives according to $\theta$.
  \item Execute one batched splatting pass with \emph{Binocular Batching} to simultaneously render left and right eye images $[L,R]$.
  \item JPEG-compress the two rendered eye buffers and package them with \texttt{frame\_id}.
  \item Stream the compressed response directly back to the HMD for display.
  
\end{enumerate}

~\\
\textbf{Implementation Details.\space\space}
GA Backend is implemented in Python~3.9\footnote{\url{https://www.python.org/}}, 
PyTorch~2.3.0\footnote{\url{https://pytorch.org/}}, and 
PyTorch3D\footnote{\url{https://pytorch3d.org/}}. 
Rendering runs on a single NVIDIA RTX~4090 GPU in real time.

%% file: sections_revised/4_Experiments.tex
\section{Experiments}

\subsection{Baselines}
To evaluate our system, we compare against mesh-based full-body avatar reconstruction methods using the SMPL-X~\cite{SMPL-X} model. \autoref{fig:baseline} illustrates the baseline systems constructed for comparison: (A) video-based mesh avatar reconstruction, and (B) image-based mesh avatar reconstruction. 
For 3DGS avatar generation in our system, VRGaussianAvatar (\textit{Ours}), a frontal image of each subject in an A-pose with arms slightly extended was captured. 
This single image served as the input for generating a full-body 3DGS avatar. 

In the case of video-based mesh avatar reconstruction (\textit{video}), we separately utilized head and body videos. For the head, we recorded a one-minute video of the Facial Action Coding System (FACS)~\cite{FACS}-based facial expressions and reconstructed it with FATE~\cite{zhang2025fate} to obtain a textured FLAME~\cite{flame} model. For the body, we captured RGBD videos using the Azure Kinect DK\footnote{\url{https://learn.microsoft.com/en-us/azure/kinect-dk/}} and reconstructed the textured SMPL-X model with RC-SMPL~\cite{rcsmpl}. The head and body textures were subsequently merged using the method of~\cite{song2025fasttexturetransferxr}. 

For image-based mesh avatar reconstruction (\textit{image}), we used the same single input image employed in VRGaussianAvatar. We adopted the off-the-shelf SMPL-X inference model SMPLer-X~\cite{cai2023smpler} for mesh recovery, and generated textures with TexDreamer~\cite{liu2024texdreamer}, an image-based SMPL-X texture generator. 



~\\
\textbf{Experimental Details. \space}
All baselines and the VRGaussianAvatar runtime were executed on a desktop PC equipped with an Intel Core i9-14900K CPU, an NVIDIA RTX~4090 GPU, and 64~GB of RAM.
For avatar reconstruction, both the image-based mesh baseline and our system’s Gaussian Avatar Module were run on the same desktop PC as the runtime environment. 
The image-based baseline required approximately 30 seconds for SMPL-X parameter estimation and 19 seconds for texture generation. 
Our Gaussian Avatar Module reconstructed a full-body 3D Gaussian Avatar within 10 seconds. 
For the video-based mesh baseline, body reconstruction with RC-SMPL took about 30 seconds, and avatar integration required an additional 2 seconds. 
Facial reconstruction with FATE was performed on a separate workstation equipped with an NVIDIA A40 GPU and required about 12 hours.
\begin{figure}[t!]
\centering
 \includegraphics[width=\columnwidth]{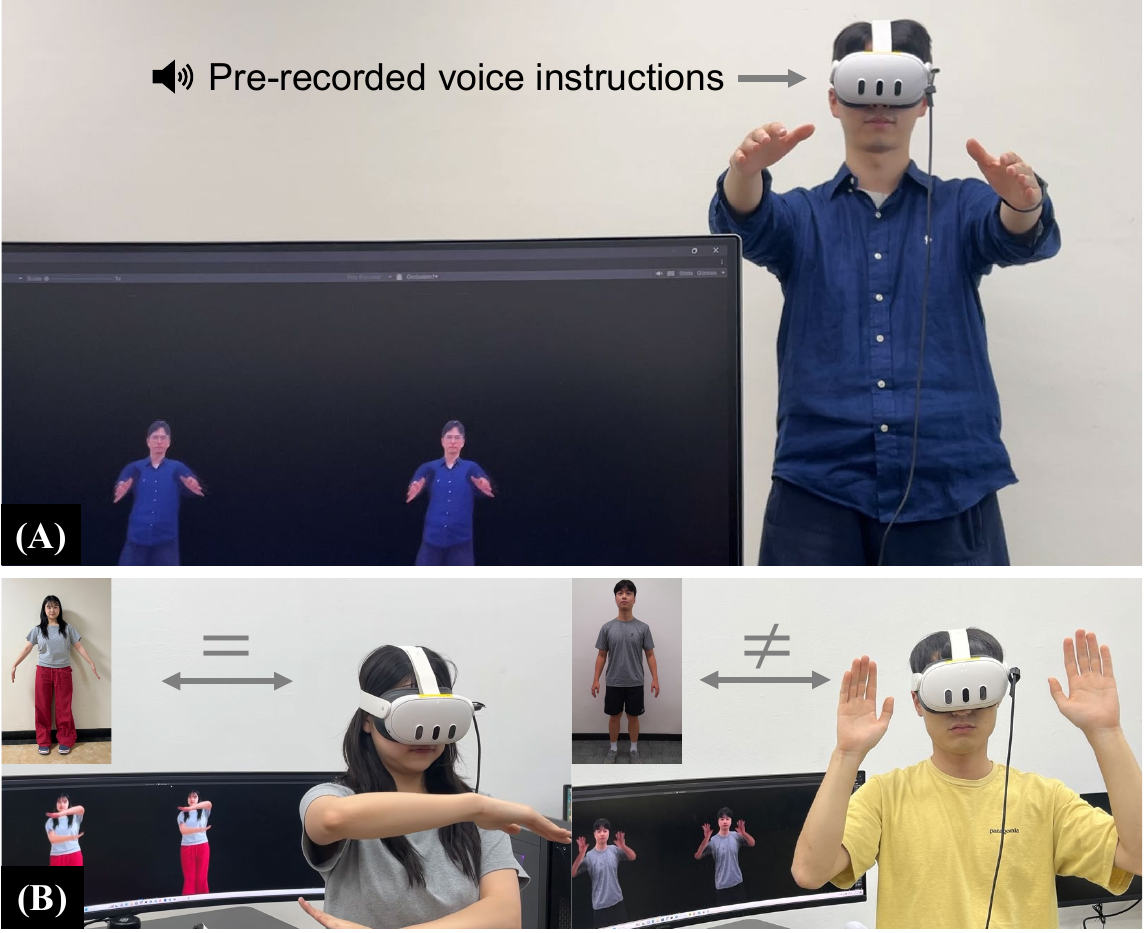}
\caption{(A) User study setup with participants performing poses from pre-recorded voice instructions. 
(B) Avatar type factor with two levels: \textit{self} (left, resembling the participant) and \textit{other} (right, resembling another person).}

\label{fig:userstudy_env}
\end{figure}

\subsection{User Study}

We conducted a human-subject study to evaluate whether our system can effectively generate realistic user avatars for VR. The evaluation considered two perspectives: (1) whether a self-resembling avatar is perceived as one’s own, and (2) whether an avatar resembling another person can be accepted as one’s own. Based on these perspectives, we compared our system with two baseline methods.

~\\
~\\
\textbf{Study Design and Task. \space\space}
The experimental factors and conditions were organized as follows. The first factor was avatar type with two levels: (1) self-representative avatar (\textit{self}) and (2) other-representative avatar (\textit{other}). For the \textit{other} avatar type, we used avatars of non-participant models matched in gender, ethnicity, and similar age to the participants (\autoref{fig:userstudy_env} (B)).
The second factor was avatar reconstruction method with three conditions related to the generation process: (1) \textit{Ours} (VRGaussianAvatar), (2) \textit{video} (video-based avatar reconstruction)~\cite{rcsmpl, zhang2025fate, song2025fasttexturetransferxr}, and (3) \textit{image} (image-based avatar reconstruction)~\cite{liu2024texdreamer, cai2023smpler} methods. 
Both of the avatar type and method factors were within-subject factors, our experiment employed a 2$\times$3 within-subject factorial design.
To compare method conditions under the self and other perspectives, avatar type was counterbalanced across participants, and the three method conditions within each type were also counterbalanced using a Latin Square design.

To evaluate the appearance and movements of avatars in VR, we adapted a commonly used avatar control task from prior studies assessing virtual embodiment~\cite{roth2020construction, do2024stepping, waltemate2018impact}.
In this task, participants wearing an HMD experienced their avatars in mirror mode and were instructed through audio cues to make specific body poses, while observing how these poses were reflected in their avatar~\cite{roth2020construction}.
We chose the mirror-based observation (rather than a purely first-person view) because our primary goal was to let participants directly inspect the reconstructed avatar’s overall appearance and motion quality as a self-representation.
Building on the previous work ~\cite{do2024stepping}, which extended the original task ~\cite{roth2020construction} by adding side-view observation instructions, we designed the task to better fit our study context and to ensure that participants could sufficiently observe the avatars. For instance, participants were encouraged to freely explore their avatars (e.g., \textit{``Move your arms and take any posture you want.''}) and were also given additional instructions to observe their avatars not only from the front, but also from the left, right, and rear views. 

~\\
\textbf{Dependent Variables. \space\space\space}
In this study, the Sense of Embodiment (SoE) was measured as the primary dependent variable, as the study aimed to evaluate whether our realistic avatars’ appearance and movements conveyed appropriate body representation and perception ~\cite{benford1995user, roth2020construction}. 
Based on the previous studies that explored overall perception and embodiment of avatars in VR~\cite{mal2023impact, wolf2022exploring, menzel2025avatars, mal20242d, waltemate2018impact}, we adopted two representative measurements: (1) the Virtual Embodiment Questionnaire (VEQ) by Roth and Latoschik~\cite{roth2020construction} and (2) VEQ+ by Fiedler et al.~\cite{fiedler2023embodiment}.
The VEQ, which is specifically designed to assess SoE in VR environments, consists of three latent factors. First, Virtual Body Ownership (VBO) refers to the perception of a virtual body as one’s own. Second, Agency (AG) denotes the perception of control over the virtual body, and the last Change (CH) captures alterations in the perceived body schema. 
The VEQ+ extends this framework by incorporating two additional dimensions: Self-location (the feeling of being located at the position of one’s body) 
and Self-identification (related to personalization and resemblance). Under these dimensions, VEQ+ has three subscales---Self-Location (SL), Self-Attribution (SA), and Self-Similarity (SS). 
Both the VEQ and VEQ+ comprise 12 items, for a total of 24 items.

The Virtual Human Plausibility Questionnaire (VHPQ) measures the extent to which a virtual avatar’s appearance and behavior are perceived as plausible, coherent, and consistent with the VR environment~\cite{mal2022virtual}. Considering the objective and setup of our system evaluation, we only used the Appearance and Behavior Plausibility (ABP) subscale of the VHPQ, which consisted of six items. 
In addition, following earlier research~\cite{roth2020construction}, we also included three self-reported items to more directly compare the realism, naturalness, and visuomotor synchrony of avatar motion behavior: \textit{``The virtual avatar’s movements were realistic,''} \textit{``The virtual avatar’s movements were naturalistic,''} and \textit{``The virtual avatar’s movements were in synchrony with my own movements.''} All questionnaire measures were assessed on a 7-point Likert scale (1: Strongly disagree – 7: Strongly agree).
For the post-experiment interview, we designed open-ended questions with reference to Menzel et al.~\cite{menzel2025avatars} to collect qualitative feedback on avatar fidelity, suitability, and realism when observing both \textit{self} and \textit{other} avatar types. Participants were also asked about their subjective opinions on differences between avatars and their preferences. 

~\\
\textbf{Procedure. \space\space\space}
This study was approved by the Institutional Review Board (IRB). Each participant attended two sessions. The first session was for data acquisition for the \textit{video} method, where participants’ facial data were scanned (pre-processing required over 12 hours); They also received an explanation of the study, provided informed consent, and completed a demographic questionnaire. The second session, conducted at least one day later, was the main experiment. At the beginning, participants’ bodies were scanned to generate avatars for the \textit{Ours} and \textit{image} conditions, while the \textit{video} avatars were generated by combining the pre-processed facial model with the newly scanned body.

After avatar generation, participants wore an HMD and performed the avatar control task while standing in an empty experimental space (\autoref{fig:userstudy_env} (A)). Following pre-recorded voice instructions, they assumed various body poses and observed their virtual avatars mirroring their poses in real-time. After each method condition, they removed the HMD and completed the post-task questionnaire. Participants first completed the three method conditions (\textit{Ours, video, image}) arranged in a balanced order under one avatar type, and then repeated the sequence under the other type, resulting in six evaluations in total. Finally, a post-experiment interview was conducted to collect qualitative feedback on their experiences with each method under both \textit{self} and \textit{other} conditions.

~\\
\textbf{Participants. \space\space\space}
We recruited 24 participants (12 male, 12 female) through the campus website. Their ages ranged from 22 to 34 years ($M =$ 27.67, $SD =$ 3.17). 
Regarding prior experience with avatar-mediated applications (e.g., social networking services, games), 17 participants reported using such applications fewer than five times (70.83\%), whereas seven participants reported using them five times or more (29.17\%). In contrast, most participants (20, 83.3\%) indicated that they had used AR/VR technologies and wearable devices more than five times, while only four participants reported fewer than five times of AR/VR experience (16.67\%).

\subsection{Quantitative Experiments}



We conducted a quantitative evaluation of the Gaussian Avatar Module, a core component of VRGaussianAvatar responsible for avatar representation. For comparison, we employed video-based~\cite{rcsmpl, zhang2025fate, song2025fasttexturetransferxr} and image-based~\cite{liu2024texdreamer, cai2023smpler} avatar reconstruction methods that are also applicable in VR, consistent with our user study setup.
We also measured the rendering speed at different resolutions to assess the contribution of Binocular Batching for real-time rendering.

~\\
\textbf{Datasets.\space\space\space}
We recruited 24 participants (12 male and 12 female) for evaluation. For both VRGaussianAvatar and the image-based baseline, a single source image was captured with the subject standing in an A-pose (arms lowered and slightly apart from the body). For the video-based baseline, we recorded videos for facial and body reconstruction. Facial reconstruction used a one-minute RGB video containing diverse expressions based on the FACS~\cite{FACS}, while body reconstruction relied on a 30-second RGBD video capturing both the frontal and rear views of the subject. For fair evaluation, rendered results from each method were cropped to the full-body and compared against the corresponding ground-truth images.

~\\
\textbf{Metrics.\space\space}
We employed four widely used metrics for evaluating avatar reconstruction quality: L1 distance, LPIPS~\cite{lpips}, PSNR~\cite{psnr}, and SSIM~\cite{ssim}.
L1 distance measures pixel-wise absolute error between the rendered image and ground truth.
LPIPS (Learned Perceptual Image Patch Similarity) assesses perceptual similarity based on deep neural network features, better reflecting human visual perception.
PSNR (Peak Signal-to-Noise Ratio) quantifies reconstruction quality in terms of signal fidelity, with higher values indicating less distortion.
SSIM (Structural Similarity Index) evaluates structural consistency and perceptual quality, particularly in terms of luminance, contrast and structure.

%% file: sections_revised/5_Results.tex
\section{Results}

\begin{figure*}[ht!]
\centering
 \includegraphics[width=\textwidth]{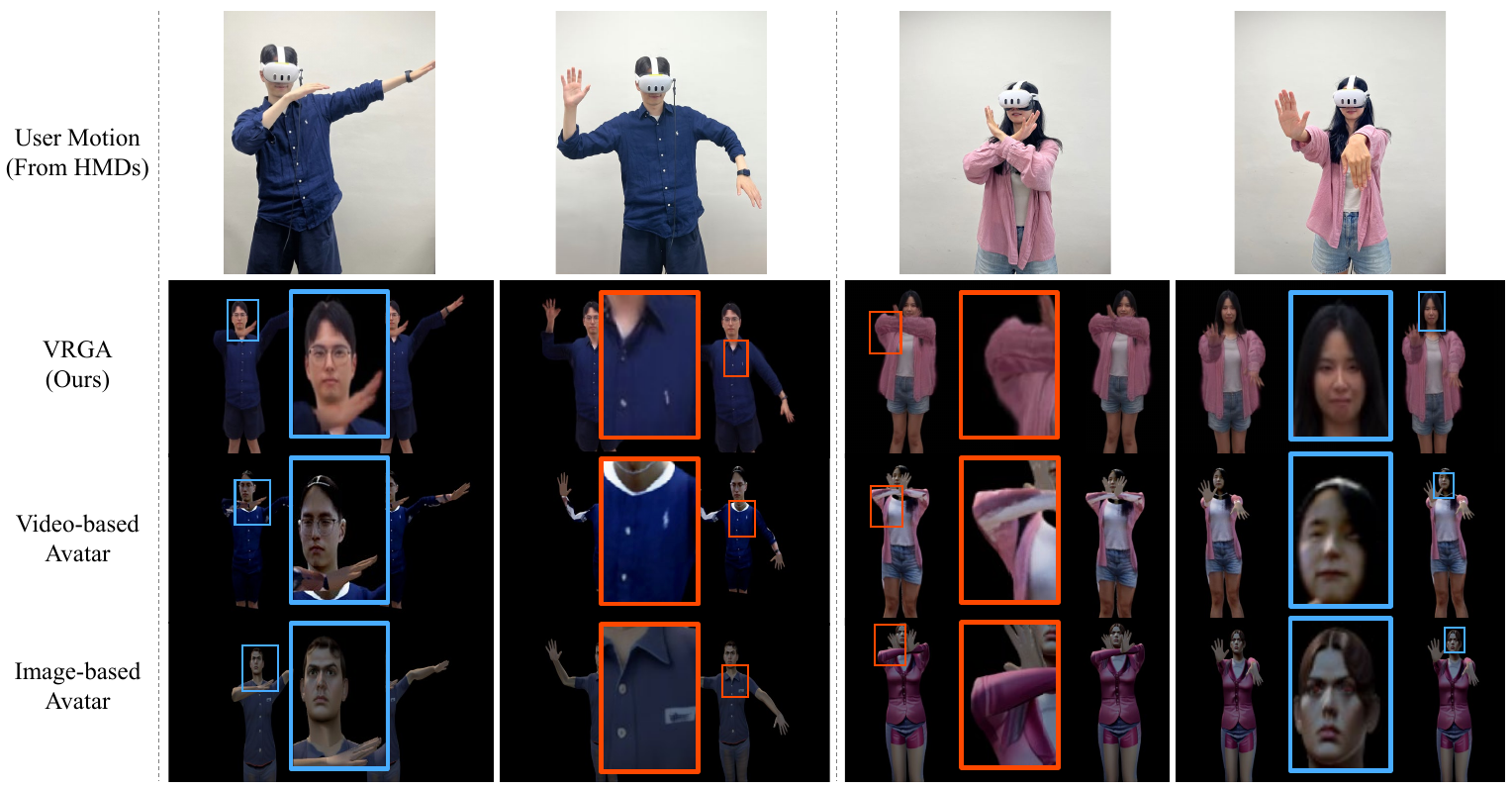}
\caption{
Qualitative comparison of VRGaussianAvatar with baseline avatar representations under identical HMD-only control.
Compared to baselines, VRGaussianAvatar typically preserves sharper appearance details in close-up regions, including head/face (sky-blue boxes) and clothing (orange boxes).
All methods are rendered stereoscopically in real time, enabling interactive avatar control in VR.
}

 \label{fig:results_1}
\end{figure*}

\subsection{User Study Analysis}

\begin{figure*}[th!]
\centering
 \includegraphics[width=\textwidth]{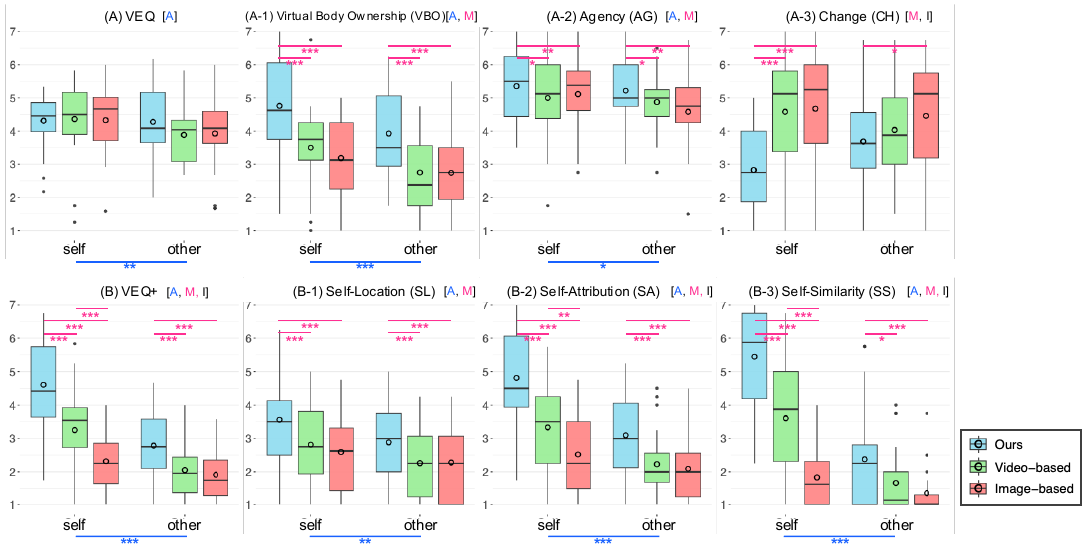}
 
\caption{Results for (A)--(A-3) Virtual Embodiment Questionnaire and its subscales; (B)--(B-3) VEQ+ and its subscales. (A and M: significant main effect of avatar type and method, respectively; I: significant interaction effect between avatar type and method)}
 \label{fig:user_graph1}
\end{figure*}

\begin{figure*}[th!]
\centering
 \includegraphics[width=\textwidth]{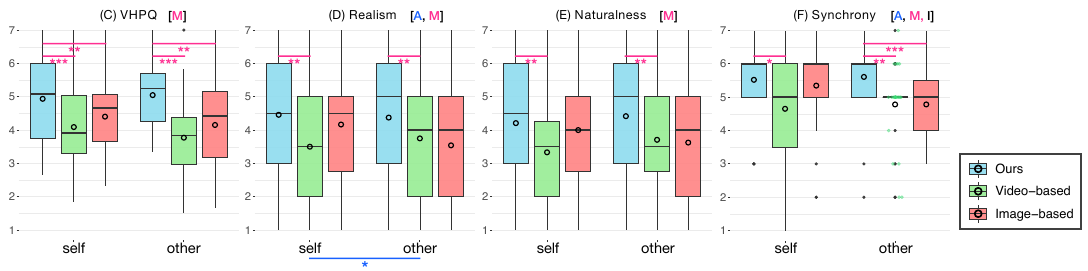}
 
\caption{Results for (C) Virtual Human Plausibility Questionnaire; (D)--(F) Additional subjective ratings. In (F), green jittered points indicate individual responses for the video-based method in the other avatar type, shown to complement the box plot where the interquartile range collapsed to a single value. (A and M: significant main effect of avatar type and method, respectively; I: significant interaction effect between avatar type and method)}
 \label{fig:user_graph2}
\end{figure*}

For subjective measures, a two-way repeated measures ANOVA with the Aligned Rank Transform (ART) was used for non-parametric factorial analysis ($\alpha =$ .05)~\cite{wobbrock2011aligned}. The ART method has been widely utilized in human-subject research analysis because it is specialized in multi-factorial non-parametric analysis and accommodates ordinal and continuous variables~\cite{wobbrock2011aligned}. All post-hoc comparisons were conducted using the ART-C procedure~\cite{elkin2021aligned} with Bonferroni corrected p-values. The internal consistency among Likert items was examined by the Cronbach’s alpha reliability coefficient. 

\subsubsection{Virtual Embodiment}
\textbf{VEQ.\space\space} 
The results for the VEQ and its subscales are shown in \autoref{fig:user_graph1}. 
The VEQ score, which aggregates all three subscales, showed a reliable consistency ($\alpha_{VEQ} =$ .825), and the three subscales also showed an acceptable value ($\alpha_{VBO} =$ .861; $\alpha_{AG} =$ .837; $\alpha_{CH} =$ .890). 
We found a main effect of avatar type ($F_{1,23} =$ 8.464, $p=$ .008, $\eta_{p}^{2} =$ .269), with \textit{self} rated higher than \textit{other} avatar types. There were no significant main effect of method ($F_{2,46} =$ 1.082, $p=$ .348, $\eta_{p}^{2} =$ .045), nor interaction effect between the two factors ($F_{2,46} =$ 2.493, $p=$ .094, $\eta_{p}^{2} =$ .098).

For VBO, significant main effects were found for avatar type ($F_{1,23} =$ 15.064, $p<$ .001, $\eta_{p}^{2} =$ .396) and method ($F_{2,46} =$ 21.903, $p<$ .001, $\eta_{p}^{2} =$ .488). Post-hoc tests indicated that \textit{self} avatars were rated higher than \textit{other}, and \textit{Ours} method induced higher ownership than both \textit{video} and \textit{image} (all $p<$ .001), while \textit{video} and \textit{image} did not differ. No interaction effect was observed ($F_{2,46} =$ 1.419, $p=$ .252).
AG showed significant main effects of avatar type ($F_{1,23} =$ 4.323, $p=$ .049, $\eta_{p}^{2} =$ .158) and method ($F_{2,46} =$ 6.041, $p=$ .005, $\eta_{p}^{2} =$ .208). Post-hoc comparisons revealed that \textit{self} avatars scored higher than \textit{other} avatars, and that \textit{Ours} conveyed higher agency than both \textit{video} ($p=$ .031) and \textit{image} ($p=$ .006), while \textit{video} and \textit{image} were not different. The interaction was not significant ($F_{2,46} =$ 0.841, $p=$ .438). 
The CH score, with lower values reflecting less perceived change from one’s own body and thus a better representation of the actual user, showed a significant main effect of method ($F_{2,46} =$ 13.728, $p<$ .001, $\eta_{p}^{2} =$ .374), but no main effect of avatar type. Pairwise comparisons showed that \textit{Ours} yielded lower CH scores than both video ($p=$ .001) and image ($p<$ .001), whereas video and image did not differ. In addition, a significant interaction effect was observed between two factors ($F_{2,46} =$ 7.932, $p=$ .001, $\eta_{p}^{2} =$ .256). Follow-up analyses showed that in the \textit{self} condition, \textit{Ours} produced lower CH scores than both \textit{video} and \textit{image} (all $p<$ .001), while in the \textit{other} condition, \textit{Ours} was only lower than \textit{image} ($p=$ .024). 

\noindent\textbf{VEQ+. } 
VEQ+ ($\alpha_{VEQ+} =$ .937) and its subscales all showed an acceptable Cronbach's alpha ($\alpha_{SL} =$ .765; $\alpha_{SA} =$ .889; $\alpha_{SS} =$ .965). The results of VEQ+ are also in \autoref{fig:user_graph1}. 
The overall VEQ+ score revealed significant main effects of avatar type ($F_{1,23} =$ 57.715, $p<$ .001, $\eta_{p}^{2} =$ .715) and method ($F_{2,46} =$ 49.487, $p<$ .001, $\eta_{p}^{2} =$ .683), and a significant interaction effect ($F_{2,46} =$ 16.155, $p<$ .001, $\eta_{p}^{2} =$ .413). Post-hoc tests showed that \textit{self} avatars scored higher than \textit{other}, and \textit{Ours} was significantly higher than both \textit{video} and \textit{image} (all $p<$ .001). Also, \textit{video} was rated higher than \textit{image} ($p<$ .001). Within the \textit{self} condition, \textit{Ours} was significantly higher than both \textit{video} and \textit{image} (all $p<$ .001), and \textit{video} was rated higher than \textit{image} ($p=$ .005). Within the \textit{other} condition, \textit{Ours} again scored significantly higher than \textit{video} and \textit{image} ($p<$ .001), while the difference between \textit{video} and \textit{image} was not significant. 

SL showed significant main effects of avatar type ( $F_{1,23} =$ 9.389, $p=$ .005, $\eta_{p}^{2} =$ .290) and method ($F_{2,46} =$ 12.948, $p<$ .001, $\eta_{p}^{2} =$ .360). Post-hoc comparisons indicated that \textit{self} avatars were rated higher than \textit{other} avatars, and \textit{Ours} showed significantly higher SL than \textit{video} ($p=$ .001) and \textit{image} ($p<$ .001), whereas \textit{video} and \textit{image} did not differ. No interaction effect was found. 
SA revealed significant main effects of avatar type ($F_{1,23} =$ 45.904, $p<$ .001, $\eta_{p}^{2} =$ .666) and method ($F_{2,46} =$ 44.525, $p<$ .001, $\eta_{p}^{2} =$ .659), as well as a significant interaction effect ($F_{2,46} =$ 10.653, $p<$ .001, $\eta_{p}^{2} =$ .317). Post-hoc analysis showed that \textit{self} type was higher than \textit{other} type; \textit{Ours} method was rated significantly higher than \textit{video} and \textit{image} (all $p<$ .001), and \textit{video} was also higher than \textit{image} ($p=$ .006). In both the \textit{self} and \textit{other} conditions, \textit{Ours} induced significantly higher scores than \textit{video} and \textit{image} (all $p\leq$ .001), and \textit{video} was higher than \textit{image} but only in the \textit{self} condition ($p=$ .005). 
Similarly, analysis of SS revealed significant main effects of avatar type ($F_{1,23} =$ 71.296, $p<$ .001, $\eta_{p}^{2} =$ .756), method ($F_{2,46} =$ 63.800, $p<$ .001, $\eta_{p}^{2} =$ .735), and their interaction  ($F_{2,46} =$ 25.390, $p<$ .001, $\eta_{p}^{2} =$ .525). Post-hoc comparisons first indicated that \textit{self} type was higher than \textit{other} type. Moreover, \textit{Ours} had significantly higher SS than \textit{video} and \textit{image}, and \textit{video} was also higher than \textit{image} (all $p<$ .001). Within the \textit{self} type, \textit{Ours} perceived more similar with the participants than both \textit{video} and \textit{image}, and \textit{video} was rated higher than \textit{image} (all $p<$ .001). In the \textit{other} type, \textit{Ours} also scored higher similarity than \textit{video} ($p=$ .027) and \textit{image} ($p<$ .001), while \textit{video} and \textit{image} had no difference. 

\subsubsection{Virtual Avatar Plausibility and Motion Behavior}
\textbf{VHPQ. } 
The results regarding the VHPQ and self-reported measures are in \autoref{fig:user_graph2}. With respect to the ABP in VHPQ, it showed a reliable Cronbach's alpha ($\alpha_{VHPQ} =$ .900). A main effect of method was observed ($F_{2,46} =$ 13.126, $p<$ .001, $\eta_{p}^{2} =$ .363), with \textit{Ours} induced higher appearance and behavior plausibility than \textit{video} ($p=$ .001) and \textit{image} ($p=$ .005). No effects of avatar type or interaction were found.

~\\ 
\textbf{Self-reported Motion Behavior. } 
For motion realism, significant main effects of avatar type ($F_{1,23} =$ 4.656, $p=$ .042, $\eta_{p}^{2} =$ .168) and method ($F_{2,46} =$ 5.244, $p=$ .009, $\eta_{p}^{2} =$ .186) were observed, while the interaction effect was not significant. 
The pairwise post-hoc tests indicated that \textit{self} avatars were higher than \textit{other} avatars, and \textit{Ours} induced higher behavior realism than video ($p=$ .009). No other differences were significant. 
Regarding naturalness, a significant main effect of method was only found ($F_{2,46} =$ 4.950, $p=$ .011, $\eta_{p}^{2} =$ .177). Similar with the realism, post-hoc analysis revealed that \textit{Ours} had significantly higher naturalness than \textit{video} ($p=$ .009), while image did not differ from either. No effect of avatar type or interaction was found. 
Lastly, one participant’s data contained an error and was excluded from the synchrony analysis. Synchrony showed a main effect of method ($F_{2,44} =$ 10.604, $p<$ .001, $\eta_{p}^{2} =$ .325), and an interaction with avatar type ($F_{2,44} =$ 3.459, $p=$ .040, $\eta_{p}^{2} =$ .136). In post-hoc analysis, it was revealed that \textit{Ours} was rated as having higher motion synchrony than both \textit{video} and \textit{image} (all $p<$ .001), and also \textit{video} was higher than \textit{image}  ($p=$ .012). In the \textit{self} condition, \textit{Ours} method had higher values than \textit{video}  ($p=$ .031). In the \textit{other} condition, \textit{Ours} was higher than both \textit{video} ($p=$ .002) and \textit{image} ($p<$ .001), while other pairs were not different.

\subsection{Quantitative Results}
\label{sec:5.2}
\begin{table}[t]
\centering
\renewcommand{\arraystretch}{1.6} 
\caption{Quantitative comparison of rendering quality across different conditions. 
Video-based avatars are constructed following~\cite{rcsmpl, zhang2025fate, song2025fasttexturetransferxr}, 
and image-based avatars are constructed following~\cite{liu2024texdreamer, cai2023smpler}.}
\begin{tabular}{lcccc}
\hline
Condition & \textbf{L1} $\downarrow$ & \textbf{LPIPS} $\downarrow$ & \textbf{PSNR} $\uparrow$ & \textbf{SSIM} $\uparrow$ \\
\hline
Video-based Avatar & 0.042 & 0.099 & 16.33 & 0.927 \\
Image-based Avatar & 0.043 & 0.103 & 16.74 & 0.917 \\
\textbf{VRGA (Ours)} & \textbf{0.033} & \textbf{0.082} & \textbf{17.92} & \textbf{0.934} \\
\hline
\end{tabular}
\label{table:quantitative}
\end{table}

\autoref{table:quantitative}  summarizes the quantitative evaluation across different conditions. 
Our method (VRGA) achieves the best performance in all metrics, 
demonstrating consistent advantages over both baselines. 
Between the two baselines, the video-based avatar shows better results than the image-based avatar 
in terms of L1, LPIPS, and SSIM.

\begin{table}[t]
\centering
\renewcommand{\arraystretch}{1.3}
\caption{
Rendering performance across different resolutions with and without Binocular Batching (BB). 
Binocular Batching enables efficient real-time stereoscopic rendering. 
Resolution indicates the per-eye resolution, and FPS is measured under binocular rendering.
}
\begin{tabular}{lc|cccc}
\hline
Resolution & \makecell{Binocular\\Batching} & \textbf{LPIPS} $\downarrow$ & \makecell{\textbf{Render}\\\textbf{Time}} & \textbf{FPS} $\uparrow$ \\
\hline
\multirow{2}{*}{512$\times$512} 
  & w/ BB  & \multirow{2}{*}{0.085} & 21.78 ms & 45.91 \\
  & w/o BB &                        & 37.47 ms & 26.69 \\
\hline
\multirow{2}{*}{768$\times$768} 
  & w/ BB  & \multirow{2}{*}{0.084} & 23.19 ms & 43.12 \\
  & w/o BB &                        & 38.33 ms & 26.09 \\
\hline
\multirow{2}{*}{1024$\times$1024} 
  & w/ BB  & \multirow{2}{*}{\textbf{0.082}} & 25.45 ms & 39.29 \\
  & w/o BB &                        & 38.80 ms & 25.77 \\
\hline
\end{tabular}

\label{table:bb_performance}
\end{table}
\autoref{table:bb_performance} presents image similarity, rendering time, and FPS across different resolutions with and without Binocular Batching (BB). 
By exploiting the fact that stereoscopic rendering in VR shares SMPL-X parameters and dynamic 3DGS attributes, our system renders both camera views simultaneously with a batch size of two, enabling high-quality avatar rendering in real time.
For the user study, we employed Binocular Batching at a resolution of 1024$\times$1024, which provided real-time performance and visual quality.

\autoref{fig:results_1} provides representative qualitative examples that visually corroborate these quantitative improvements, showing clearer appearance details for VRGaussianAvatar under identical HMD-only control. More qualitative results are provided in Appendix~\ref{app:qualitative}.

%% file: sections_revised/6_Conclusion.tex
\section{Discussion}
\label{sec:discussion}
The goal of this work is to enable practical, high-fidelity avatar representation for VR applications by addressing the limitations of existing video- and image-based approaches. To this end, we introduce VRGaussianAvatar, an HMD-only controllable avatar pipeline that integrates 3D Gaussian Splatting into an interactive VR system, and evaluate whether these technical contributions translate into perceptually meaningful improvements through a user study.

Our method produced stronger virtual embodiment than video and image-based approaches, particularly in body ownership and agency. Participants reported that the avatar looked and moved like their own body, with less alteration and greater resemblance. While aggregated VEQ scores did not differ significantly, subscales revealed that the sense of ``my body'' was more clearly enhanced with our method (\textit{``felt the closest to how I actually move,''} P7).
Further analyses of VEQ+ showed notable advantages in self-location and self-identification (self-attribution, and self-similarity). Participants described feeling that their body shifted into the avatar’s position in VR and that the avatar both resembled and responded to them. Even in other-representative conditions, where similarity scores were relatively low, our method still surpassed alternatives by producing more human-like and relatable avatars. By contrast, video-based avatars were described as \textit{``overall texture seemed reflected but with unnatural movement,''} and image-based as \textit{``only reflecting clothing color, not my actual features''} (P2).

Beyond embodiment, our system achieved the highest plausibility and realism, with more natural and synchronized movements than video-based avatars. Although some participants acknowledged that image-based avatars behaved smoothly or without latency (P13, P18), they emphasized that our method offered more consistent control and stronger human-likeness(\textit{``moved naturally and felt alive,''} P15).
Finally, these advantages translated into higher preference and practical usability. Most participants (79.2\%) selected our method as the avatar they would most prefer and want to use in a virtual space, describing it as \textit{``resembled me the most and felt natural'' }(P20, P23). Even when evaluating ``other'' avatars, Ours was the most appealing to participants (50\%) while maintaining higher plausibility and synchrony, underscoring its applicability for social VR scenarios. 
Taken together, these findings show that our method better represents virtual embodiment, identification, and realism than baseline approaches, providing a solid foundation for both self and other-representative avatars in VR.

These results indicate that the perceptual advantages observed in the user study may be associated with the technical design of the proposed pipeline. By separating real-time pose estimation from high-cost Gaussian rendering through the VR Frontend and GA Backend architecture, the system maintains stable and responsive control while preserving visual fidelity, which may help support users’ sense of embodiment and agency. In addition, Binocular Batching enables consistent stereoscopic rendering without redundant computation, allowing 3DGS-based avatars to remain interactive and perceptually plausible.
As such, VRGaussianAvatar provides a practical foundation for deploying high-fidelity, controllable avatars in scenarios where both visual realism and embodied interaction play central roles; it is expected to show particularly high applicability in multi-user immersive environments in which users must continuously perceive both themselves and others as coherent agents (e.g., social VR, remote collaboration, or games). While further validation with larger samples and extended tasks is needed, the ability to support both self- and other-representative avatars within a single, efficient pipeline has the potential to broaden its applicability.

\subsection{Limitations and Future Work}
\label{sec:limitations}
Our system has several limitations.
First, because we do not integrate facial sensing signals, the avatar does not support dynamic facial expressions, gaze, or blinking.
Second, our reconstruction assumes a controlled input image that is favorable for one-shot full-body reconstruction; performance may degrade under substantial occlusions or poses that deviate from the capture guideline.
Third, while GPU-side rendering runs in real time on a desktop GPU, the end-to-end system latency is subject to streaming overhead. While our implementation runs on localhost, deploying the GA Backend over a remote network would increase end-to-end latency and may require bitrate--quality trade-offs (e.g., lower resolution or stronger compression) to maintain interactivity.
Finally, our user study primarily relies on mirror-based observation and therefore does not directly evaluate embodiment outcomes from an egocentric (first-person) viewpoint.


Future work will aim to improve practicality and generalizability by addressing the limitations above.
First, for social VR scenarios, we plan to support dynamic facial control by incorporating head-avatar reconstruction/control techniques (e.g., VOODOO XP~\cite{tran2024voodooxp}).
Second, to reduce reliance on controlled capture, we will explore robustness strategies for more unconstrained inputs.
Third, to move toward standalone deployment, we will investigate on-device avatar representation by adopting lightweight 3DGS avatar techniques (e.g., SqueezeMe~\cite{iandola2025squeezeme}).
Fourth, we will conduct additional user studies that directly evaluate embodiment from a first-person viewpoint.

\subsection{Ethical Considerations}
\label{sec:ethics}

Because our system can generate high-fidelity avatars from a single image, it may be misused for non-consensual digital cloning or impersonation.
To mitigate such risks, we restrict the integrated system to research and educational use and require written consent from the subject prior to avatar creation, including a clear specification of the intended use and duration.
In addition, XR deployments that present reconstructed avatars should explicitly disclose the driving user to reduce deception risks.
Our user study was conducted under Institutional Review Board (IRB) approval and followed standard consent and privacy procedures.
If considering broader deployment, safeguards will likely require a combination of technical and procedural measures, such as consent protocols, identity verification, and standardized watermarking.

\section{Conclusion}
Building on these findings, we presented a system for real-time full-body 3DGS avatar representation in VR environments that can be controlled solely through internal sensors of commercial HMDs.
Our framework integrates a VR Frontend, which estimates controllable full-body motion from these sensor signals, with a GA Backend that performs efficient stereoscopic rendering via Binocular Batching. 
This parallel design enables photorealistic and responsive avatar experiences without the need for additional external trackers, lowering the barrier for practical VR deployment.
In addition, our quantitative evaluations and user studies confirm that VRGaussianAvatar not only achieves more photorealistic avatar representation compared to mesh-based avatar systems reconstructed from images and videos, but also provides higher levels of virtual embodiment and plausibility for users.

In summary, VRGaussianAvatar demonstrates that 3D Gaussian Splatting can be effectively extended to real-time, photorealistic, and HMD-only controllable avatar representation in VR. We believe this work highlights the feasibility of real-time 3DGS-based avatars and opens the door to richer and more engaging social interaction in immersive environments.

%% file: sections_revised/7_Appendix.tex
\clearpage
\appendix

\twocolumn[{%
\vspace*{0.6cm}
\begin{center}
    {\LARGE \textbf{VRGaussianAvatar: Integrating 3D Gaussian Splatting Avatars into VR}}\\[6pt]
    {\Large Appendix.}
\end{center}
\vspace*{0.4cm}
}]

\section{Implementation Details}
\label{sec:appendix_implementation}

\subsection{VR System Configuration}
\label{subsec:vr_system_config}
The VR runtime consists of a Unity-based VR Frontend and a Python-based GA Backend running on the same local machine. The HMD is connected via a wired link to ensure stable high-bandwidth transmission. Key system parameters are summarized in~\autoref{tab:system_params}.
\begin{table}[h]
\centering
\caption{VR System and Rendering Hyperparameters.}
\label{tab:system_params}
\begin{tabular}{l|c}
\toprule
\textbf{Parameter} & \textbf{Value} \\
\midrule
\multicolumn{2}{c}{\textit{Rendering Backend (Python)}} \\
\midrule
Render Resolution (per eye) & $1024 \times 1024$ \\
Field of View (FOV) & $65.0^\circ$ \\
Inter-pupillary Distance (IPD) & 0.055 m \\
JPEG Compression Quality & 80 \\
Binocular Batching & Enabled (Single Pass) \\
\midrule
\multicolumn{2}{c}{\textit{VR Frontend (Unity)}} \\
\midrule
Target HMD & Meta Quest 3 \\
Connection Type & Wired Link (USB 3.0) \\
Network Protocol & WebSocket (Localhost) \\
IK Solver & Final IK (VRIK) \\
Update Rate & Client-driven (Async) \\
\bottomrule
\end{tabular}
\end{table}

\subsection{Offline Reconstruction Pipeline in GA Backend}
\label{subsec:lhm_details}
We utilize the Large Human Model (LHM-500M) ~\cite{qiu2025lhm} architecture for the single-image 3D Gaussian Splatting reconstruction. This architecture leverages powerful pre-trained encoders, specifically DINOv2 and Sapiens, to extract semantic and geometric features from a single RGB image, which are then decoded into 3D Gaussian parameters via a transformer-based network. For a more detailed description of the reconstruction pipeline and training strategy, please refer to LHM~\cite{qiu2025lhm}. The detailed hyperparameters for the model architecture and training process are listed in ~\autoref{tab:lhm_hyperparams}.
\begin{table}[h]
\centering
\caption{Hyperparameters for the Offline Reconstruction Pipeline.}
\label{tab:lhm_hyperparams}
\resizebox{0.95\linewidth}{!}{%
\begin{tabular}{l|l|c}
\toprule
\textbf{Category} & \textbf{Parameter} & \textbf{Value} \\
\midrule
\multirow{5}{*}{Model Architecture} 
 & Image Encoder & DINOv2-ViT-L/14 (Frozen) \\
 & Fine Encoder & Sapiens-1B (Frozen) \\
 & Transformer Config & 5 Layers, 16 Heads, 1024 Hidden Dim \\
 & Point Embeddings & 1024 Dimensions \\
 & Gaussian MLP & 2 Layers, 512 Neurons, SiLU Activation \\
\midrule
\multirow{5}{*}{Training} 
 & Optimizer & AdamW ($\beta_1=0.9, \beta_2=0.95$) \\
 & Learning Rate & $4 \times 10^{-5}$ (Cosine Schedule) \\
 & Weight Decay & 0.05 \\
 & Batch Size & 4 (per GPU) \\
 & Total Epochs & 60 \\
\midrule
\multirow{4}{*}{Loss Weights} 
 & Masked Pixel Loss & 1.0 \\
 & Perceptual Loss & 1.0 \\
 & Face ID Loss & 0.05 \\
 & Offset Regularization & 1000.0 \\
\bottomrule
\end{tabular}%
}
\end{table}

\subsection{Streaming Protocol and Latency}
\label{subsec:streaming_latency}
While our reference setup uses localhost, deploying the GA Backend over a remote network can introduce additional latency and bandwidth constraints.
Maintaining interactive frame delivery in such settings may therefore require bitrate--quality trade-offs (e.g., stronger compression or lower resolution), depending on network conditions.

We implement a WebSocket-based streaming protocol in the reference setup.
The VR Frontend sends per-frame HMD pose and SMPL-X parameters as JSON, and the GA Backend returns rendered stereo images as concatenated JPEG binaries.
\autoref{tab:latency_breakdown} reports the end-to-end latency breakdown measured under this localhost configuration.


\begin{table*}[t]
\centering
\caption{
Latency breakdown of the streaming pipeline measured under the localhost configuration.
Stages are grouped by where they execute (GA Backend vs.\ VR Frontend).
}
\label{tab:latency_breakdown}
\begin{tabular}{l l c}
\toprule
\textbf{Stage} & \textbf{Location} & \textbf{Time (ms)} \\
\midrule
Tracking \& inverse kinematics & VR Frontend & $\sim 1$ \\
\midrule
Localhost communication (request/response) & GA Backend $\leftrightarrow$ VR Frontend & $< 1$ \\
\midrule
Stereo rendering (Binocular Batching) & GA Backend & $\sim 23\text{-}27$ \\
Image encoding (JPEG, multi-threaded) & GA Backend & $\sim 1\text{-}2$ \\
\midrule
Image decoding \& display & VR Frontend & $\sim 5\text{-}10$ \\
\midrule
\textbf{End-to-end latency} & -- & $\mathbf{\sim 30\text{-}41}$ \\
\bottomrule
\end{tabular}
\end{table*}

Note that these stages run asynchronously in a pipelined manner.
As a result, the achievable frame rate is primarily bounded by the GA Backend rendering time (and encoding), while the end-to-end latency reflects the full pipeline delay from tracking to display.
We report FPS in \autoref{table:bb_performance} using the GA Backend throughput (rendering + encoding).

\section{More Qualitative Results and Challenging Cases}
\label{app:qualitative}

\begin{figure*}[ht!]
\centering
 \includegraphics[width=\textwidth]{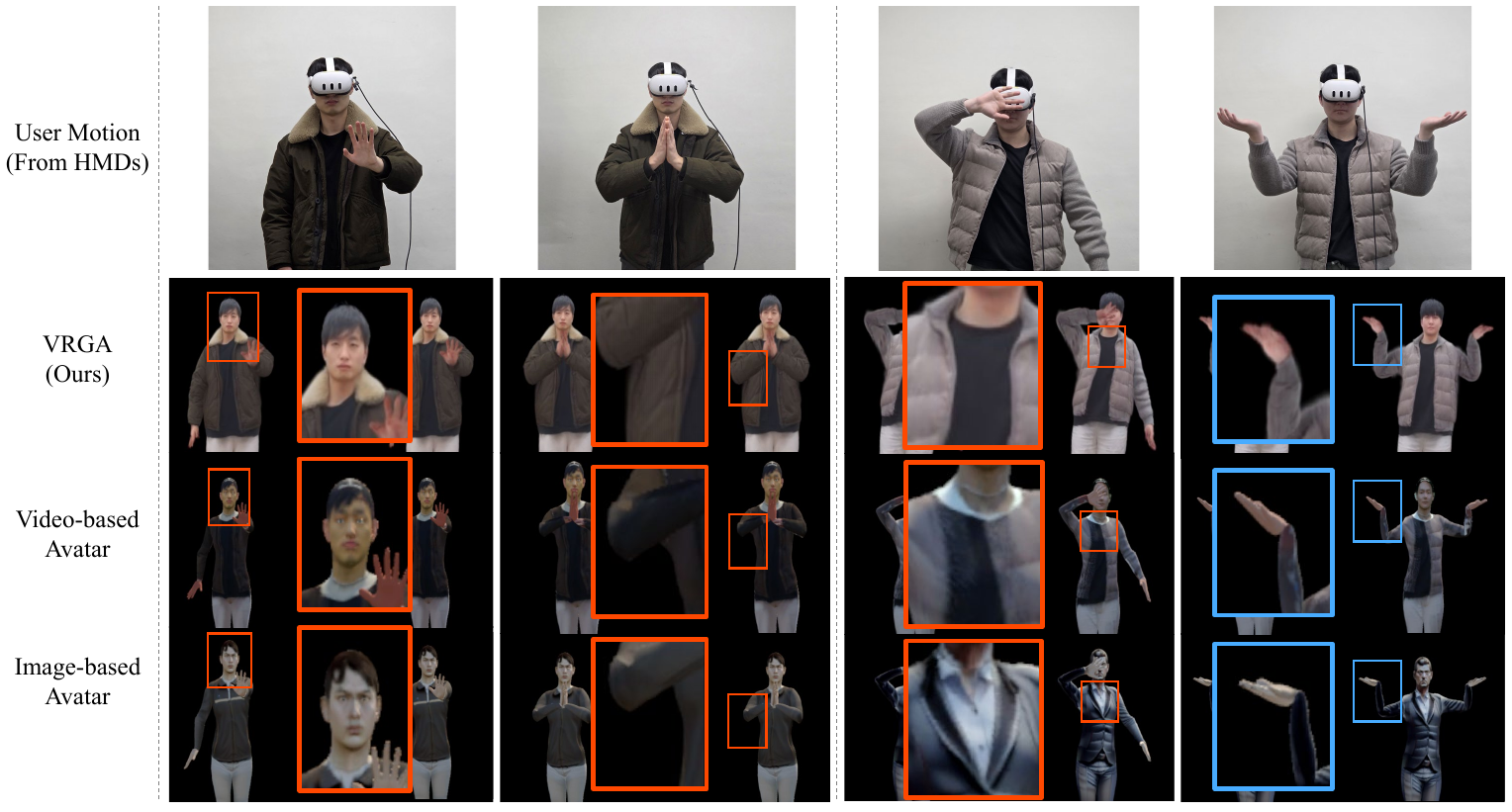}
\caption{
\textbf{More qualitative results.}
Orange boxes indicate representative close-ups where our method preserves facial and clothing details.
Sky-blue boxes highlight a challenging hand region, discussed in \autoref{app:qualitative}.
}

 \label{fig:results_2}
\end{figure*}

This section provides additional qualitative comparisons of VRGaussianAvatar against
baseline mesh avatar representations under identical HMD-only control.
We highlight representative regions where our 3DGS avatar preserves fine appearance cues
(e.g., facial details and clothing texture), and we include a single challenging case
focused on hands.
Representative examples are shown in \autoref{fig:results_2}.

Hands are particularly difficult to render sharply in our current pipeline.
Compared to other body regions, hands have a higher kinematic freedom and undergo larger
pose changes during natural interaction.
However, the underlying 3DGS hand representation is reconstructed with a finite Gaussian
density, which can be insufficient to stably represent fast, highly articulated
deformations.
As a result, our method can exhibit localized blurring around fingers during motion,
and this artifact can be more noticeable than in relatively rigid regions or in
mesh-based reconstructions.

As future work, we plan to incorporate a dynamic hand pose representation and integrate
a VR-oriented 3DGS hand model that is designed for frequent, highly articulated hand motions.

